# Modeling State-Conditional Observation Distribution using Weighted Stereo Samples for Factorial Speech Processing Models

Mahdi Khademian[1] · Mohammad Mehdi Homayounpour[1]

**Abstract** This paper investigates the effectiveness of factorial speech processing models in noise-robust automatic speech recognition tasks. For this purpose, the paper proposes an idealistic approach for modeling state-conditional observation distribution of factorial models based on weighted stereo samples. This approach is an extension to previous single pass retraining for ideal model compensation which is extended here to support multiple audio sources. Non-stationary noises can be considered as one of these audio sources with multiple states. Experiments of this paper over the set A of the Aurora 2 dataset show that recognition performance can be improved by this consideration. The improvement is significant in low signal to noise energy conditions, up to 4% absolute word recognition accuracy. In addition to the power of the proposed method in accurate representation of state-conditional observation distribution, it has an important advantage over previous methods by providing the opportunity to independently select feature spaces for both source and corrupted features. This opens a new window for seeking better feature spaces appropriate for noisy speech, independent from clean speech features.

**Keywords** factorial speech processing models, state-conditional observation distribution, stereo samples, noise-robust automatic speech recognition

✉ Mohammad Mehdi Homayounpour
 homayoun@aut.ac.ir

 Mahdi Khademian
 khademian@aut.ac.ir

[1] Laboratory for Intelligent Multimedia Processing (LIMP), Amirkabir University of Technology, Tehran, Iran

# 1. Introduction

Despite long term efforts and great successes in automatic speech recognition (ASR) systems, rapid degradation of performance in the presence of noise and other competing sources remains the Achilles Heel of these systems [2]. While some feature enhancement and model adaptation techniques loosely use noise source characteristics to increase the performance of speech recognition systems, model based methods try to incorporate as much as information they can acquire from noise sources [11]. Such information ranges from statistics of stationary noises to dynamic state transition patterns of cyclo-stationary noises.

Factorial speech processing models [7, 17, 20] are extensions of Hidden Markov Models (HMM) which model audio sources and the way that these sources are combined in a generative manner. They model each of the audio sources separately; this can include modeling dynamic changes of audio sources by hidden Markov models. Additionally, factorial models model how these audio sources are combined to produce output or distorted features. For this reason, factorial models can incorporate more details of noise characteristics for improving robust-ASR system performance.

Figure 1 shows a generic HMM for conventional acoustic modeling in speech recognition and a factorial model for noise robust speech recognition. Models are expressed in the Probabilistic Graphical Models (PGM) language. The depicted HMM is extended in two ways for creating a factorial model of speech processing. First, factorial models have multiple state chains which are useful for modeling systems with multiple underlying independent processes [3], i.e. two audio sources with their corresponding temporal state changes (in Fig. 1.b $s_t^n$ and $s_t^x$ are noise and speech source state variables). These multiple chains with their observation models construct source models of factorial models (see Fig. 1.b). Gaussian Mixture Models are usually used for representing the observation models. Increasing the number of underlying Markov chains increases the computational requirements of inference, exponentially. This is known as one of the challenges in factorial models [12]. The second extension is the interaction model, [11]; the distribution of the observed signal feature conditioned on features of its corresponding sources. In Fig. 1.b CPD of $p(y_t|x_t, n_t)$ represents it. This CPD in its deterministic form is called mismatch function [8, 22]; $p(y_t|x_t, n_t) = \delta(y_t - f(x_t, n_t))$.

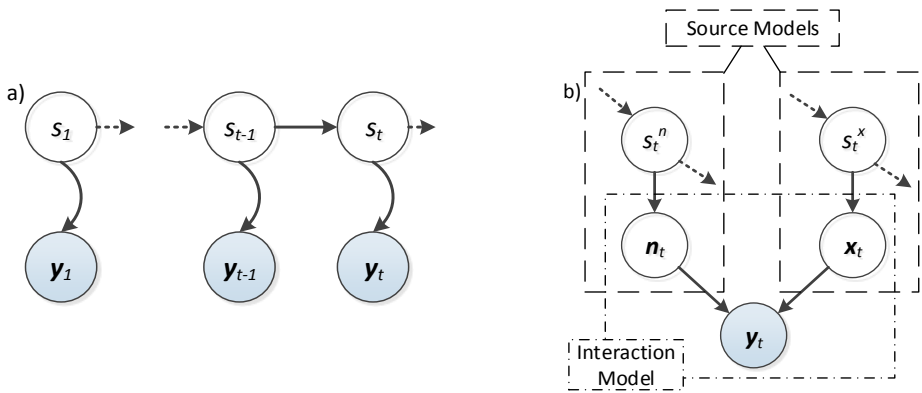

**Fig. 1.** a) A generic HMM for acoustic modeling, b) Factorial speech processing model for robust-ASR

The interaction model is not directly used in the inference. Inference on factorial models requires State-Conditional Observation Distribution (SCOD); i.e. observed feature distribution conditioned on states of the source chains, $p(\mathbf{y}_t|s^x, s^n)$. This distribution is directly calculated by marginalizing-out source feature variables, i.e. $\iint p(\mathbf{y}, \mathbf{x}, \mathbf{n}|s^x, s^n)d\mathbf{x}d\mathbf{n}$. The direct calculation of SCOD causes more challenges in use of factorial models which necessitates additional approximations. As a result, having more accurate interaction models leads to more approximations to the SCOD calculation. On the other hand, doing exact SCOD calculations forces us to use approximate interaction models.

The current paper focuses on the second challenge of factorial models by incorporating the idea of single-pass retraining [8, 24] for ideal SCOD representation. In addition, it presents a modified expectation maximization (EM) algorithm for parametric modeling of this ideal SCOD. In fact, the procedure behind the proposed method for SCOD modeling inherently resolves the need for using mismatch function and direct calculation of the SCOD. Therefore, none of the two approximations mentioned above are involved in the created models by the proposed method. This increases the accuracy of the created SCOD models.

Next section briefly reviews previous methods for modeling the SCOD by starting with the description of the commonly used environment model and the most applicable mismatch function for speech recognition. Its last subsection presents probabilistic inference of factorial models of speech processing. In addition, it discusses the computational complexity of inference in these models. The proposed method of SCOD modeling is proposed in section 3; in this section, an extension of the EM algorithm for parametric modeling of SCODs is described. Section 4 describes experiments by providing a block diagram of the proposed method and implementation details for different test scenarios; Aurora 2 dataset is used for this evaluation. Section 5 presents the evaluation results and finally the last section concludes the paper.

We believe that unified view of this paper for SCOD modeling based on the past model compensation techniques in addition to its proposed method provides valuable insight into the factorial models of speech processing.

## 2. Background

In model based noise-robust ASR methods, the following relation is considered between speech and noise signals in an assumed environment for generation of distorted speech signals [5, 7]:

$$y = x * h + n \qquad (1)$$

where $x$ and $n$ are speech and noise signals and $h$ is the channel model of the recording environment. In the power spectrum domain after framing and windowing by short-term discrete Fourier transform we have:

$$|Y_t|^2 = |X_t H|^2 + |N_t|^2 + 2|X_t H||N_t| \cos(\boldsymbol{\phi}_t) \qquad (2)$$

in which $\boldsymbol{\phi}_t$ is the vector of phase difference between frequency bins of $\boldsymbol{X}_t\boldsymbol{H}$ and $\boldsymbol{N}_t$ complex vectors. In addition, $\boldsymbol{X}_t$ and $\boldsymbol{N}_t$ are extracted from the frame $t$ of their corresponding time signals. In (2), frequency index is omitted since variables are written in the vector form; it will be provided in subsequent relations where needed. Time index will be removed in subsequent expressions for brevity.

While (2) provides a mismatch function useful for speech enhancement applications, power spectrum is not appropriate for speech recognition. By applying filterbank, the following relation is extracted for filterbank energies of sources and corrupted signals (this derivation is based on an approximation in which the channel model is considered to have a flat frequency response for each filter in the filterbank) [22]:

$$\bar{Y}_i = \bar{X}_i\bar{H}_i + \bar{N}_i + 2\alpha_i\sqrt{\bar{X}_i\bar{H}_i\bar{N}_i} \tag{3}$$

where $\bar{Y}_i$ denotes weighted averaged power spectrum energy obtained from the $i$th filter as:

$$\bar{Y}_i = \sum_k w_{ik}|Y_k|^2 \tag{4}$$

where $k$ is the frequency bin index and $w_{ik}$ is the $i$th filter weights across different frequency bins, $k$. Filterbank energies for clean speech and noise frames are also calculated similar to distorted speech; i.e. $\bar{X}_i = \sum_k w_{ik}|X_k|^2$, $\bar{N}_i = \sum_k w_{ik}|N_k|^2$.

In (3), $\alpha_i$ is called the phase factor which reflects effect of phase difference between the sources, averaged by the $i$th filter in different frequency bins which equals to:

$$\alpha_i = \sum_k w_{ik}|X_k||H_k||N_k|\cos(\phi_k)/\sqrt{\bar{X}_i\bar{N}_i} \tag{5}$$

By considering uniform distribution for phase difference, $\alpha_i$ becomes a stochastic variable whose support set is [-1, 1]; more of its properties is investigated in [14]. Using the logarithm and truncated discrete cosine transform matrix (DCT), the following interaction model is derived for MFCC features:

$$p(\boldsymbol{y}^c|\boldsymbol{x}^c,\boldsymbol{h}^c,\boldsymbol{n}^c) = \delta\left(\boldsymbol{y}^c - \boldsymbol{C}\log\left(\exp(\boldsymbol{C}^{-1}(\boldsymbol{x}^c+\boldsymbol{h}^c)) + \exp(\boldsymbol{C}^{-1}\boldsymbol{n}^c) + \boldsymbol{\epsilon}(\boldsymbol{x}^c,\boldsymbol{h}^c,\boldsymbol{n}^c)\right)\right) \tag{6}$$

where its residual equals to:

$$\boldsymbol{\epsilon}(\boldsymbol{x},\boldsymbol{h},\boldsymbol{n}) = 2\boldsymbol{\alpha}\exp(\boldsymbol{C}^{-1}(\boldsymbol{x}+\boldsymbol{h}+\boldsymbol{n})/2) \tag{7}$$

Interaction models useful for speech recognition applications contain this residual. The interaction model is usually approximated by removing the residual [15] or considering its phase factor as constant which has the same value across all frequency bins [16]. To the best of our knowledge, almost all model compensation techniques were developed based on these interaction models by accepting the mentioned approximations. The proposed method of this work resolves the need for using mismatch function since its SCOD model is not directly calculated by the interaction models. Therefore its derived SCOD models are not extracted by several approximations involved in developing interaction models.

## 2.1. State-conditional observation distribution

As mentioned earlier, the interaction model is not directly used in inference of factorial models. Instead, by marginalizing source features as in (8), the state-conditional observation distribution is calculated for inference.

$$p(y|s^x, s^n) = \iint p(y, x, n|s^x, s^n) dx dn = \iint p(y|x, n) p(x|s^x) p(n|s^n) dx dn \quad (8)$$

Direct calculation of (8) is performed for raw feature domains by considering some approximations in mismatch function. Two examples are max and soft-max approximation for log-power-spectral features [11, 19]. But this domain is not appropriate for speech recognition. Therefore, three categories of approaches are used for SCOD modeling. Approaches in the first set, by making an assumption that the SCOD is Gaussian in specific feature domains, estimate Gaussian parameters. Parallel model combination (PMC) is an example of this group [9]. Methods in the second set, approximate the non-linear mismatch function by linearizing it around an expansion point using Taylor series; then they transform source model parameters by applying this approximation. Several variations of these methods were developed so far. A successful and complete vector Taylor series (VTS) based compensation method is presented in [16]. The third set uses conditional samples of observation distribution for estimating parameters of the SCOD model which is usually modeled by Gaussian Mixture Models (GMM). Samples are generated by forward sampling and consequent use of mismatch function. Developed methods in this set are known as variations of data-driven PMC [8] (DPMC).

Next sub-sections describe VTS and DPMC based SCOD modeling adopted for factorial speech processing models. Most of the previous works support only stationary noises and therefore consider one noise state in their SCOD models. Actually, in this case, clean source models are replaced with their corresponding SCODs; it means replacing observation models of the original HMM. This is the reason for naming this kind of robust speech recognition, "model compensation" [22]. However this work considers noises with multiple states and re-state previous methods, supporting non-stationary noises.

### 2.1.1 VTS based SCOD models

In the VTS based methods, mismatch function is approximated by the first order Taylor series expanded around source mean vectors, $(x_0 = \mu_x, h_0 = \mu_h, n_0 = \mu_n)$. Therefore for mismatch function of (6) in the form of $y = f(x, h, n)$, we have:

$$y \approx f(x_0, h_0, n_0) + \frac{\partial f}{\partial x}(x - x_0) + \frac{\partial f}{\partial h}(h - h_0) + \frac{\partial f}{\partial n}(n - n_0) \quad (9)$$

The above linear approximation transforms source model Gaussians into the corrupted feature space. For each state of speech and noise, we have the following SCOD by selecting their corresponding mean vectors as the expansion point:

$$p(y|s^x = i, s^n = j) \sim \mathcal{N}(y; \mu_i + \mu_h + g(\mu_i, \mu_h, \mu_j), G\Sigma_i G^T + F\Sigma_j F^T) \quad (10)$$

in which $g(x, h, n) = C \log \left( \exp(C^{-1}(x+h)) + \exp(C^{-1}n) + \epsilon(x, h, n) \right)$, $G = \frac{\partial f}{\partial x}$, and $F = \frac{\partial f}{\partial n}$. For source models with GMM observation distribution, the SCOD can be conditioned on each source component distribution. Additionally for delta and delta-delta coefficient the SCOD parameters are extracted as follows:

$$p(y_\Delta | s^x = i, s^n = j) \sim \mathcal{N}(y_\Delta; G\mu_{\Delta i} + F\mu_{\Delta j}, G\Sigma_{\Delta i}G^T + F\Sigma_{\Delta j}F^T) \quad (11)$$

$$p(y_{\Delta\Delta} | s^x = i, s^n = j) \sim \mathcal{N}(y_{\Delta\Delta}; G\mu_{\Delta\Delta i} + F\mu_{\Delta\Delta j}, G\Sigma_{\Delta\Delta i}G^T + F\Sigma_{\Delta\Delta j}F^T) \quad (12)$$

Detail derivation of these expressions for single state noise models can be found in [16, 22]; extending them for multiple noise states is straightforward.

### 2.1.2 DPMC based SCOD models

In data-driven parallel model combination (DPMC) methods, by using forward sampling, state-conditional observed feature samples are extracted to be used for SCOD modeling. First, source states are fixed; i.e. $\langle s^x = i, s^n = j \rangle$. Then based on the fixed states and use of source models, conditional source features are generated. Now by use of an appropriate mismatch function, observed features are extracted from the source features ($y_{l|i,j} = f(x_{l|i,j}, n_{l|i,j})$). These samples can represent an empirical SCOD as follows:

$$p(y | s^x = i, s^n = j) = \frac{1}{L} \sum_{l=1}^{L} \delta(y_{l|i,j} - y) \quad (13)$$

where $L$ is the number of samples and $\delta$ is Dirac delta function. At this step, parametric model of SCOD can be trained using state-conditional samples. This model may consist of single Gaussian or multiple Gaussians where the method is named DPMC and iterative-DPMC (IDPMC), respectively [8, 22].

### 2.2. Inference

In factorial models with multiple hidden Markov chains such as models that are used in multi-talker speech recognition and robust speech recognition tasks, the objective of inference is to find the most probable source states given the observation feature vectors, i.e.:

$$s_{1:T}^{*x,n} = \underset{s_{1:T}^{x,n}}{\mathrm{argmax}}\, p(s_{1:T}^{x,n} | y_{1:T}) \quad (14)$$

where in the noise robust speech recognition tasks, noise states are discarded since only speech states are used in the recognition.

Finding the most probable states conditioned on observation vectors is done by a two-dimensional Viterbi search. Naïve implementation requires likelihood evaluation of $(S^x S^n)^T$ different paths among source states. By creating a mega-state HMM from factorial HMM, the number of operations reduces to $O(T(S^x S^n)^2)$. In mega-state HMM, a new state variable is defined by Cartesian product of the source states. Thus, similar to HMM, decoding requires $O(TS^2)$ operations, where $S = S^x S^n$. But by using a two-dimensional Viterbi search, this reduces to $O(TS^{2x}S^n)$. The following recursions are used in a two-dimensional Viterbi search to find the most probable speech states:

$$\tau_t(s_t^x, s_{t+1}^n) = \max_{s_t^n} P(s_{t+1}^n|s_t^n)p(y_t|s_t^x, s_t^n)\tau_{t-1}(s_t^x, s_t^n) \tag{15}$$

$$\tau_t(s_{t+1}^x, s_{t+1}^n) = \max_{s_t^x} P(s_{t+1}^x|s_t^x) \tau_t(s_t^x, s_{t+1}^n) \tag{16}$$

in which $P(s_{t+1}^n|s_t^n)$ and $P(s_{t+1}^x|s_t^x)$ are transition probabilities in source Markov chains and $p(y_t|s_t^x, s_t^n)$ is the SCOD. Recursions start with:

$$\tau_0(s_1^x, s_1^n) = P(s_1^x)P(s_1^n) \tag{17}$$

Similar to conventional Viterbi algorithm, in each step, a back-pointer is used to determine target state sequence:

$$\phi_t(s_t^x, s_{t+1}^n) = \underset{s_t^n}{\mathrm{argmax}}\, P(s_{t+1}^n|s_t^n)p(y_t|s_t^x, s_t^n)\tau_{t-1}(s_t^x, s_t^n) \tag{18}$$

$$\phi_t(s_{t+1}^x, s_{t+1}^n) = \underset{s_t^x}{\mathrm{argmax}}\, P(s_{t+1}^x|s_t^x) \tau_t(s_t^x, s_{t+1}^n) \tag{19}$$

For more details on the two chain models or general cases of algorithm, the reader is referred to [12] or [3, 10] respectively. In fact, in this case, dynamic programming is run also within time-slices in addition to the standard Viterbi algorithm which only runs between time-slices. This is the reason for the reduction of computations by a factor of $S^n$.

## 3. The proposed method

The proposed method in this paper solves the problem of modeling state-conditional observation distribution in a way different from previous methods. In our method, there is no use of interaction function, therefore this method is not limited to enforcements of the interaction functions such as accordance of feature spaces. The procedure for parametric SCOD modeling is described in the next three subsections.

### 3.1. Sampling corrupted features

Sampling of corrupted features is started from source signals in time domain. At the first step, segments of source signals are combined together by the freely assumed environment model (such as (1)) to make the corrupted speech in the time domain, $y$. Then, the corresponding features are extracted from these time domain signals, i.e. $x_l$, $n_l$ and $y_l$ are $l$th sources and corrupted features. These feature vectors are known as stereo features [1, 15]. We also call them "stereo" due to one to one mapping between these features. However, in this case, three sets of features are related together.

In the next step, source features are examined in their corresponding source models to compute their state-conditional likelihoods, ($p(x_l|i)$ and $p(n_l|j)$) for all source states. Finally, the time domain segments and source feature vectors are discarded. Samples of $y$ and their source state-conditional likelihoods will be used for the modeling later.

To summarize, as shown in Fig. 2, the input for creating corrupted feature vectors are source signals and the environment model to combine these signals. Feature vectors are then extracted from all signals, independent from each other. Source feature state-conditional likelihoods are also calculated for weighting corrupted features, later.

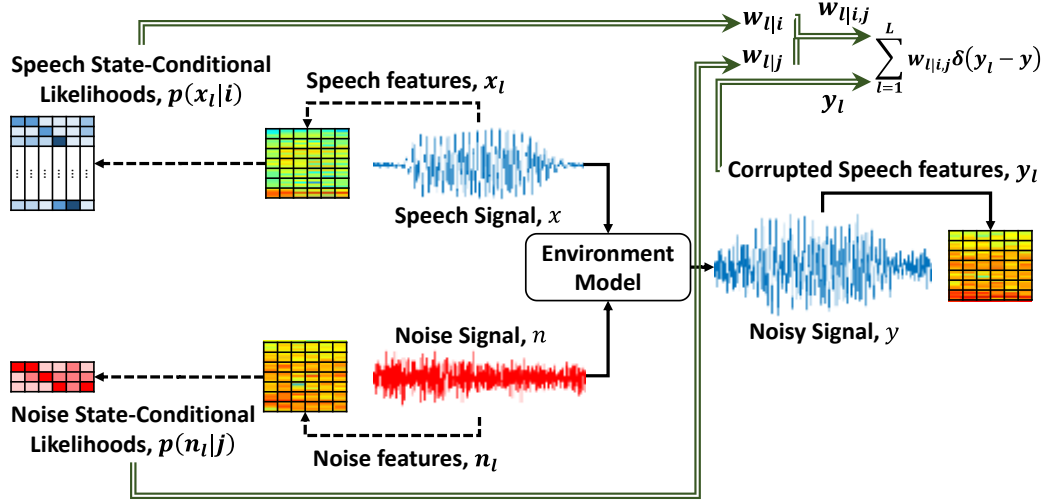

**Fig. 2.** Extraction of corrupted speech features and empirical SCOD based on the weighted features. As we can see, extracted samples using the proposed method are generated neither by forward sampling nor by use of approximated mismatch functions as in the previous data-driven methods.

### 3.2. Empirical distribution

Since in the sampling procedure there is no fixed source states, samples of $y$ are extracted from non-conditional observation distribution (the corrupted feature space). Comparing to data-driven PMC methods, these samples cannot be used directly for SCOD modeling. We use importance sampling scheme [18] to correct bias occurred by non-conditional samples for modeling the SCOD using particle weights which indicates association of particles to states. Particle weights are calculated in each source spaces as follows:

$$w_{l|i} = p(x_l|i)/p(x_l) = p(x_l|i)/\sum_i p(x_l|i)p(i) \qquad (20)$$

$$w_{l|j} = p(n_l|j)/p(n_l) = p(n_l|j)/\sum_j p(n_l|j)p(j) \qquad (21)$$

By assuming independence of the sources (which is true for many additive noise environments), we have:

$$w_{l|i,j} = w_{l|i} w_{l|j} \qquad (22)$$

Now the SCOD can be modeled empirically by the weighted particles as follows:

$$p(y|i,j) = \sum_{l=1}^{L} w_{l|i,j} \delta(y_l - y) \qquad (23)$$

where $y_l$ is the $l$th particle sampled from $p(y)$ and $w_{l|i,j}$ is its adjusting weight for $p(y|i,j)$. By iterating $i$ and $j$ through their corresponding random variable support set and calculating $w_{l|i,j}$, the empirical SCOD is extracted for the all source states. Therefore in the proposed method the SCOD is represented by a set of weighted corrupted samples, where the weights are calculated by evaluation of their corresponding stereo features in the corresponding source models. We call this procedure SCOD modeling using weighted stereo samples, WSS method which is illustrated in the Fig. 2.

Comparing WSS empirical distribution to empirical distribution of IDPMC, we can observe that while the extracted samples in IDPMC are equally weighted they are extracted using the approximated mismatch functions. In the proposed method, samples are directly extracted from the corrupted signal without any use of the mismatch function and then are weighted for modeling the observation distribution conditioned on source states. Therefore the constructed empirical distribution based on the proposed method is more close to the true SCOD comparing to other methods which are based on approximated mismatch functions. This claim will be investigated in the experiments.

Note that for the channel effect, while in the assumed environment model such as (1), $h$ remains constant which means that channel characteristics does not change quickly, the above particle weights are still applicable. For the cases where there is multiple channel states, particle weights can also be conditioned on channel state as well as source states. Deriving this extension is straightforward.

### 3.3. Parametric distribution

The empirical distribution cannot be used directly in the recognition application and parametric model of the SCOD is needed for inference. The SCOD may be modeled parametrically by single or multiple component Gaussians as in the conventional acoustic modeling. Parameter estimation of single Gaussian models can be done by maximum weighted likelihood estimators [23] as:

$$\underset{\theta}{\mathrm{argmax}} \prod_{l=1}^{L} p^{w_{l|i,j}}(\mathbf{y}_l; \theta) \tag{24}$$

Therefore Gaussian parameters are estimated by weighted samples as follows:

$$\widehat{\boldsymbol{\mu}}_{i,j} = \left(\sum_{l=1}^{L} w_{l|i,j} \mathbf{y}_l\right)/w_{i,j} \tag{25}$$

$$\widehat{\boldsymbol{\Sigma}}_{i,j} = \left(\sum_{l=1}^{L} w_{l|i,j} (\mathbf{y}_l - \widehat{\boldsymbol{\mu}}_{i,j})(\mathbf{y}_l - \widehat{\boldsymbol{\mu}}_{i,j})^T\right)/w_{i,j} \tag{26}$$

in which $w_{i,j} = \sum_{l=1}^{L} w_{l|i,j}$. Depending on feature space, single Gaussian component may not be sufficient for SCOD modeling. In these cases, GMM is used as a more flexible modeling tool. While the EM algorithm is used for training GMMs, the standard algorithm does not support weighted samples. Because of this, the algorithm is extended to support weighted samples.

Consider the following $Q$-function as the expected value of weighted complete-data log likelihood. In this $Q$-function the expectation is taken over the posterior of the latent variable for supporting weighted samples (state indices are omitted for brevity of notation):

$$Q(\theta, \theta') = \mathbb{E}_{z|y;\theta'}[\ln \mathcal{L}(\mathbf{y}_{1:L}, w_{1:L}, z_{1:L}; \theta)] = \sum_{l=1}^{L} w_l \mathbb{E}[\ln p(\mathbf{y}_l, z_l; \theta)] \tag{27}$$

For mixture of Gaussians with $\theta = (\boldsymbol{\mu}_{1:K}, \boldsymbol{\Sigma}_{1:K}, \boldsymbol{\pi})$ in which $\pi_k$ is the component's prior, the $Q$-function becomes:

$$\sum_{l=1}^{L} w_l \sum_{k=1}^{K} \gamma_l(k)[\ln p(\mathbf{y}_l; \boldsymbol{\mu}_k, \boldsymbol{\Sigma}_k) + \ln \pi_k] \tag{28}$$

where $k$ is index of Gaussian component in the mixture and $\gamma_l(k)$ is defined as the $k$th component responsibility to the $l$th sample based on old parameters ($\theta'$) as follows:

$$\gamma_l(k) \stackrel{\text{def}}{=} P(z = k|\mathbf{y}_l; \theta') \qquad (29)$$

In fact, this posterior is the outcome of the E-step of the EM algorithm (calculation of component responsibilities is provided in the appendix A). Equation (27) is weighted version of the standard EM $Q$-function for supporting weighted samples. This $Q$-function must be optimized with respect to $\theta$, new parameter set, during the M-step of the EM algorithm. Optimizing the $Q$-function with respect to new parameters leads to the following parameter update equations (detailed derivation is provided in the paper appendix A):

$$\boldsymbol{\mu}_k = (\sum_{l=1}^{L} w_l \gamma_l(k) \mathbf{y}_l)/W_k \qquad (30)$$

$$\boldsymbol{\Sigma}_k = (\sum_{l=1}^{L} w_l \gamma_l(k)(\mathbf{y}_l - \boldsymbol{\mu}_k)(\mathbf{y}_l - \boldsymbol{\mu}_k)^T)/W_k \qquad (31)$$

$$\pi_k = W_k/W \qquad (32)$$

in which $W_k = \sum_{l=1}^{L} w_l \gamma_l(k)$ and $W = \sum_{l=1}^{L} w_l$. By applying the extended EM algorithm, weighted "stereo" samples can be used to model the SCOD for all source states. Again, note that in the above formulae, source state subscripts are removed for brevity.

## 4. Experiment Setups

The Aurora 2 task [13] for recognizing utterances of digit series corrupted by additive and convolutive noises is selected for evaluating the effectiveness of the proposed method. This task has three test sets: A, B and C and our method is evaluated using set A. Set A is designed to test robustness of recognition methods against additive noises considering this point that the noise information could be used during the training phase. In this test set, four noises are artificially added to 8440 clean utterances. Corrupted utterances are used during the training phase for multi-condition training scenario and the same noises are used for creation of the test set. The four noises are Subway, Babble, Car and Exhibition which are artificially added to clean utterances in different signal to noise ratios (SNR) varying from 20 dB to -5 dB in -5 dB steps.

Before describing the details of the conducted experiments, the procedure of the proposed method for noise-robust ASR applications is described; Figure 3 shows its block diagram. Three main phases of this procedure are: source modeling, SCOD modeling and test. In the source modeling, speech and noise models are trained from clean speech utterances and noise signals. Source model parameters are state priors and transition matrices ($\pi, A$) and parameters of the GMM observation models which are Gaussian means ($\boldsymbol{\mu}_i$), covariance matrices ($\boldsymbol{\Sigma}_i$) and component weights ($M_i$).

In the next step, SCOD models are trained based on weighted "stereo" feature samples. Sampling procedure is described in section 3.1 and sample weights are calculated by (20), (21) and (22). Then the SCOD models are trained by weighted

"stereo" samples using the proposed extended EM algorithm. SCOD parameters are the same as HMM observation model parameters except that SCOD models are conditioned on both sources.

Finally, in the test phase, features of the test utterance are extracted and state-conditional observation likelihoods are calculated to perform decoding using the two-dimensional Viterbi algorithm.

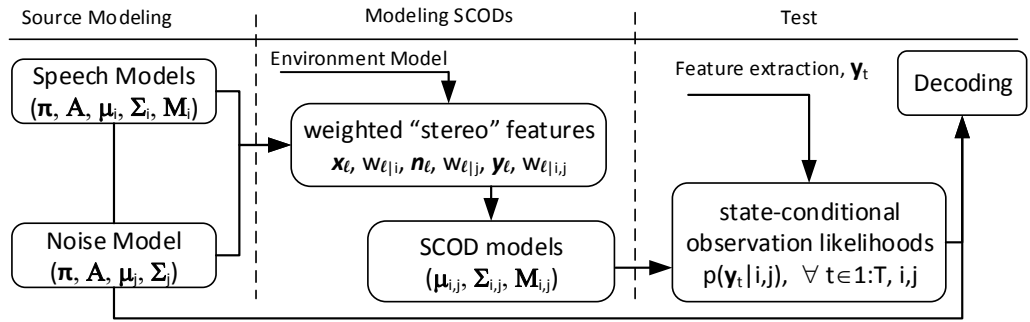

**Fig. 3.** Block diagram of the proposed method for SCOD modeling and decoding using factorial speech processing models.

### 4.1. Source modeling, Clean Speech Models

Speech source models are created using the HTK toolkit [24]. Models are trained by the Aurora 2 standard recognition scripts [13] in clean condition training mode, except that for the front-end we use Voicebox toolbox [4]. Framing and windowing are done similar to the standard Aurora 2 recognition scripts.

For comparing the performance of the ideal SCOD models trained by the WSS method to previous methods, we first limit ourselves to Mel-Frequency Cepstral Coefficients (MFCC). While the proposed method for SCOD modeling allows us independently select feature spaces of source and observed features.

Since second order derivatives of MFCCs provide no significant improvements in the experiments we only use first order derivatives. Instead of frame's logarithm of energy, zero order coefficient of MFCC features is used. In addition, the applied filterbank to power spectrum only contains 13 filters which enables us to use full DCT matrix in feature extraction and normal inverse of DCT matrix in the mismatch functions. Feature spaces of speech, noise and corrupted signals are selected to be the same with 26 coefficients (MFCC0d(26)), 13 MFCCs with their first order derivatives.

For each digit of the dataset, a sixteen state HMM model is trained using the Aurora 2 standard scripts and for silence and short pause, three and one state HMMs are used. Observation models of digits have three component GMMs while silence and short pause models use six component GMMs, all with diagonal covariance matrices.

### 4.2. Source Modeling, Noise Models

For evaluating effectiveness of factorial models in speech processing for handling non-stationary noises, two sets of noise models are used in the experiments. In the first set, each noise model has only one multivariate Gaussian. Models trained in the first set are not appropriate for non-stationary noises, since they have only one multivariate Gaussian and they have only one state.

Noise models of the second set are created by STACS tool [21] and noise modeling is done for each noise separately. This tool starts with single state model and increases HMM states to find the best model selected by the BIC criterion. Each state has a multivariate Gaussian with diagonal covariance matrix for its observation distribution. Trained models for Subway, Babble, Car, and Exhibition noises contain 3, 8, 4 and 4 states accordingly which is optimized by the STACS tool. Models of the first and second sets are used in the experiments based on VTS, IDPMC and the proposed WSS SCOD modeling technique.

### 4.3. Modeling state-conditional observation distributions

Three SCOD modeling techniques were presented in this paper and are compared together in the experiments. These include VTS and IDPMC based models and models created by the proposed method, WSS.

In the experiments, channel effect is not considered explicitly, therefore channel impulse response is removed from the environment model of (1). Consequently, channel feature vectors are removed from all of mismatch functions and the VTS based SCOD models. The reason is that in the test set A of the dataset there is no mismatch in noise environment in the training and test phases.

For the VTS based experiments, SCOD models are extracted for each Gaussian component of speech and noise states. This is done by use of (10) and (11) where the two alpha values are selected in the residual term (discussed later).

For the IDPMC based models, generated samples are extracted from source state GMMs. Then based on these conditional samples, three component block diagonal GMMs are trained for each joint states ($M = 3$).

For the proposed method about 17000 speech utterances from train set of dataset are used for creating "stereo" features. These utterances are selected randomly from the clean train set of the dataset. Random segment of the corresponding noise is selected to create corrupted utterances based on (1), ignoring the channel effect. Gain coefficient is adjusted to simulate SNRs from -5 dB to 20 dB including infinity. Voice activity detection and speech energy determination is done by the tools provided in the Voicebox [4], based on ITU recommendation P.56 (similar to the Aurora 2 test sets). Number of mixture components for modeling of SCOD is determined experimentally and is set to three for GMMs ($M = 3$) with full covariance matrices.

### 5. Results

In the first experiment, performance of compensated system (single noise state factorial model) based on IDPMC and VTS based SCOD models for two alpha values (equation (5)) are compared. Use of two selected alpha values enables us to evaluate the mentioned methods based on mismatch functions in two extreme conditions. By setting the alpha equal to zero, we ignore the effect of phase factor in interaction model of (6). The second alpha value is selected to be the same as in [16] which yields its best results. Average word recognition accuracy for four noises of set A against different SNRs for two selected alpha values are plotted in Fig. 4.

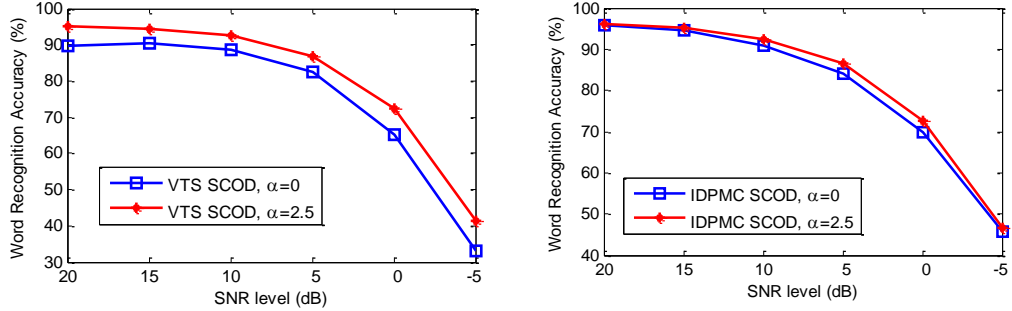

**Fig. 4.** Average recognition performance of compensated systems based on VTS and IDPMC SCOD models for different alpha values on set A of the Aurora 2 dataset over four noises.

As mentioned in [16] selecting alpha to invalid constant 2.5 (invalid regarding to its support set which is $[-1, 1]$), provides better result than ignoring it, both in VTS and IDPMC based models but with less effect in the IDPMC. Therefore the alpha value is set to 2.5 for further experiments. Additionally we observed that IDPMC based models yield higher recognition rate than VTS models.

In the next experiment the proposed method for training ideal SCOD models is compared to VTS and IDPMC methods. The experiment is done for single noise state and multiple noise states. Figure 5.a shows the comparison by average word recognition accuracy over four noises in the single noise state mode. In addition, Fig. 5.b shows absolute improvement in average recognition accuracy for these three methods when multiple noise states are used.

Moreover, Fig 5.a shows the performance of multi-conditioned trained system against three compensation approaches. This shows that even for one noise state, the performance of compensated system is greater than multi-conditioned trained system especially in low SNR conditions.

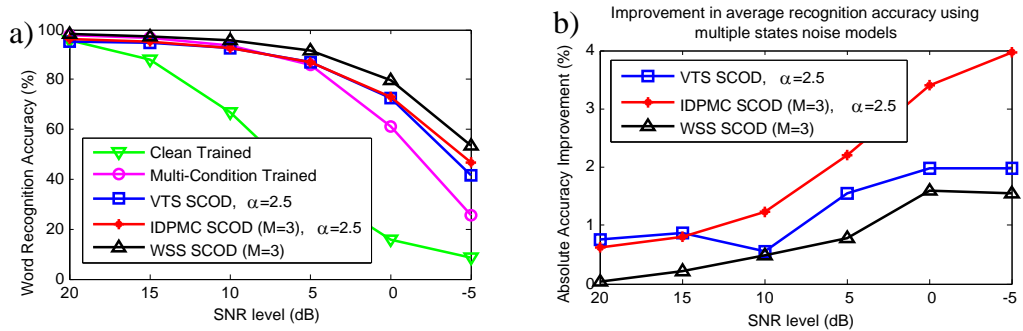

**Fig. 5.** Average recognition performance of SCOD models based on WSS, IDPMC and VTS methods on set A of the Aurora 2 dataset over four noises. For the IDPMC and WSS methods, three component Gaussian GMMs are used for SCOD modeling (M=3). a) shows average performance of single noise state models. b) shows absolute performance improvement when multiple noise states are used.

Table 1 shows detailed recognition accuracy of the experiment in the multiple noise states mode. As it can be seen, using multiple noise states improves recognition accuracy in all cases and more improvement is also achieved in the low level SNRs. Additionally while IDPMC based factorial models gain more from using multiple state noise models, WSS based factorial models achieve the best results (see table 1). This is due to the fact that "stereo" data provides frame-level mapping between the clean speech noise signal and corrupted utterances. Hence a better implicit mapping of actual corrupted features to clean speech and noise signal features is provided. This is because the other methods establish this relationship using the approximated mismatch functions.

Table 1 Detailed word recognition accuracy for set A of the Aurora 2 dataset; comparing performance of factorial model of speech processing based on multiple state noise models for different SCOD modeling methods (average is calculated over SNRs 20 to 0).

|  | VTS based SCOD models | | | | | | | IDPMC based SCOD models | | | | | | | WSS based SCOD models | | | | | | |
|---|---|---|---|---|---|---|---|---|---|---|---|---|---|---|---|---|---|---|---|---|---|
| SNR level | 20 | 15 | 10 | 5 | 0 | avg | -5 | 20 | 15 | 10 | 5 | 0 | avg | -5 | 20 | 15 | 10 | 5 | 0 | avg | -5 |
| Subway | 95.4 | 95.5 | 93.4 | 88.9 | 77.6 | **90.1** | 52.7 | 96.8 | 95.9 | 93.4 | 89.4 | 78.5 | **90.8** | 58.5 | 98.3 | 97.8 | 96.4 | 93.4 | 83.0 | **93.8** | 61.6 |
| Babble | 96.0 | 94.5 | 90.9 | 84.6 | 66.8 | **86.6** | 38.1 | 96.3 | 95.4 | 92.4 | 86.2 | 68.5 | **87.8** | 37.5 | 98.0 | 97.2 | 95.5 | 90.5 | 75.4 | **91.3** | 40.9 |
| Car | 96.4 | 95.9 | 94.4 | 90.8 | 77.0 | **90.9** | 36.4 | 96.6 | 96.1 | 94.7 | 91.1 | 80.9 | **91.9** | 53.3 | 97.8 | 97.7 | 96.9 | 94.5 | 86.2 | **94.6** | 58.9 |
| Exhibition | 96.1 | 95.9 | 94.7 | 89.4 | 75.7 | **90.3** | 46.3 | 97.2 | 96.3 | 94.5 | 88.8 | 76.7 | **90.7** | 53.6 | 98.3 | 97.7 | 96.5 | 91.1 | 81.0 | **92.9** | 58.2 |
| Average | 96.0 | 95.4 | 93.3 | 88.4 | 74.3 | **89.5** | 43.4 | 96.7 | 96.0 | 93.7 | 88.9 | 76.2 | **90.3** | 50.7 | 98.1 | 97.6 | 96.3 | 92.4 | 81.4 | **93.2** | 54.9 |

For evaluating effectiveness of factorial models for different noises, detailed relative percentage of recognition accuracy improvement of the last experiment for different noises are provided in Table 2. In fact, values provided in this table are relative improvement of WSS method for SCOD modeling for four noises when we use multiple noise states over the case where only one noise state is used. Two observations from this table can be noted. First, we can see that the factorial models are more beneficial when the corruption is more severe; i.e. low SNR conditions and more disturbing noises. Second, since non-stationary noises require multiple noise states for source modeling (see section 4.2), factorial models are more effective against noises with more states. It is apparent that for the Babble noise with 8 states, improvement is substantial and for the Subway noise with 3 states, the lowest improvement is achieved.

Table 2 Detailed percentage of word recognition relative accuracy improvement when multiple noise states are used (details for averaged improvement in Fig. 5.b for the WSS based SCOD models).

|  | 20 | 15 | 10 | 5 | 0 | avg | -5 |
|---|---|---|---|---|---|---|---|
| **Subway** | -- | 0.06 | 0.42 | -- | -- | 0.10 | -- |
| **Babble** | **0.18** | **0.47** | **0.52** | **2.85** | **5.85** | **1.97** | **11.63** |
| **Car** | 0.03 | 0.33 | **0.69** | 0.69 | 1.22 | 0.59 | 1.27 |
| **Exhibition** | -- | 0.06 | 0.45 | 0.17 | 1.43 | 0.42 | 2.56 |
| **Average** | 0.05 | 0.23 | 0.52 | 0.93 | 2.13 | 0.77 | 3.86 |

Finally, by incorporating independent feature spaces for clean and noisy speech as the advantage of the proposed method, we repeat the last experiment. Here observation features are selected to be 21 log Mel-scale filterbank energies with their first order derivatives (42 coefficients). Noise models with multiple states are used as in the last experiment. In this experiment, clean and noisy speech features are remained to be MFCC0d26. Table 3 compares the result of this experiment and the best result achieved with WSS SCOD models (right column in Table 1).

**Table 3** Comparing average word recognition accuracy over the four noises when the observation feature space is changed to LogMelFBd42.

|  | 20 | 15 | 10 | 5 | 0 | avg | -5 |
|---|---|---|---|---|---|---|---|
| LogMelFBd42 | 97.96 | 97.55 | **96.67** | **93.33** | **83.07** | **93.72** | **56.07** |
| MFCC0d26 | **98.09** | **97.62** | 96.32 | 92.38 | 81.39 | 93.16 | 54.88 |

Table 3 shows that filterbank energies are more appropriate for speech recognition in low SNR conditions which can be explained as follows. When log of filterbank energies are used as features, usually some subbands are more affected by noise due to the fact that most of noises affect only some subbands. However, when DCT is used to obtain MFCC coefficients, it uses energies of all subbands (including those affected by noise) and distributes that information to all MFCC coefficients. Therefore all MFCC feature dimensions are affected by noise corruption. As a result, it is more appropriate to use MFCC features for source modeling and features like filterbank energies for the observation space.

## 6. Conclusion

In this paper, a new method based on weighted "stereo" samples for modeling state-conditional observation distribution is proposed for use in factorial speech processing models. In fact, the idea behind this method is similar to single pass retraining technique presented in [8] for model compensation. We present this method in the context of factorial speech processing models with its support for non-stationary noises. At first, we saw that using ideal SCOD models improves system performance in model compensation scenarios. Moreover, it is shown that using multiple noise-states will increase recognition accuracy especially in low SNR conditions and for severe non-stationary noises.

Due to the use of "stereo" data, the proposed method cannot be used directly in many real applications since this data is not always available in the training phase. But similar to the ideal compensated models in the single pass retrained systems, we are able to train ideal SCOD models to assess capabilities of other practical techniques. The purpose for presenting this method is to provide a way to investigate whether increasing the number of noise states in noise models is useful for non-stationary noises in order to improve the overall system performance or not. As a result, increasing system performance in our experiments encourages researchers for developing methods for use in factorial speech processing models capable of handling non-stationary noises. The proposed method is still applicable in noise specific environments where noise information is available in advance during the training phase and in such a setting it performs far better than multi-condition trained systems.

**Acknowledgements** The authors would like to thank to Mohammad Ali Keyvanrad, Dr. Omid Naghshineh Arjmand and Dr. Adel Mohammadpour for their valuable arguments and suggestions in this work.

## Appendix A

**Extending the EM algorithm for modeling mixture of Gaussians based on weighted samples**

In the E-step of the EM algorithm for weighted particles, particle weights have no effect on the component responsibility equations. By considering particle weights as the replicating order of the particles (similar to (24)), we see that this replication has no effect on the component responsibilities to each particle. Therefore component responsibilities are calculated without considering particle weights by the old parameter set as in E-step of the standard EM algorithm for GMMs:

$$\gamma_l(k) \propto \pi'_k \mathcal{N}(y_l; \mu'_k, \Sigma'_k) \tag{33}$$

where the normalization constant is $\sum_{k=1}^{K} \pi'_k \mathcal{N}(y_l; \mu'_k, \Sigma'_k)$.

For the M-step, the following optimization problem must be solved:

$$\theta^{new} = \underset{\theta}{\operatorname{argmax}} Q(\theta, \theta') \\ st: \sum_{k=1}^{K} \pi_k = 1 \tag{34}$$

Using the method of Lagrange multiplier for satisfying the constraint for component priors, we have the following objective function for optimization:

$$g(\mu, \Sigma, \pi) = \sum_{l=1}^{L} w_l \sum_{k=1}^{K} \gamma_l(k)[\ln p(y_l; \mu_k, \Sigma_k) + \ln \pi_k] + \lambda \left(\sum_{k=1}^{K} \pi_k - 1\right) \tag{35}$$

Taking the derivative $g$ with respect to $\mu_k$ results in:

$$\partial g / \partial \mu_k = 2 \sum_{l=1}^{L} w_l \gamma_l(k) [\Sigma_k^{-1}(y_l - \mu_k)] \tag{36}$$

Now (30) is easily obtained for updating $\mu_k$ by setting this derivative to zero. For estimating $\Sigma_k$, according to [6] the derivative takes the following form:

$$\partial g / \partial \Sigma_k = -\frac{1}{2} \sum_{l=1}^{L} w_l \gamma_l(k) \left[\Sigma_k^{-1} - \Sigma_k^{-1}(y_l - \mu_k)(y_l - \mu_k)^T \Sigma_k^{-1}\right] \tag{37}$$

in which the $\mu_k$ is estimated by (30). Setting it to zero, we obtain:

$$\sum_{l=1}^{L} w_l \gamma_l(k)(y_l - \mu_k)(y_l - \mu_k)^T \Sigma_k^{-1} = \sum_{l=1}^{L} w_l \gamma_l(k) \tag{38}$$

Then (31) is obtained for estimating $\Sigma_k$ in which when the number of samples are significant, there is no need for adjusting the estimator for bias. Finally for $\pi_k$ we have:

$$\partial g / \partial \pi_k = \sum_{l=1}^{L} (w_l \gamma_l(k))/\pi_k + \lambda = 0 \tag{39}$$

by using the assumption $\sum_{k=1}^{K} \pi_k = 1$ and considering $\gamma_l(k)$ as a valid conditional probability mass function, $\lambda$ is calculated by:

$$\lambda = -\sum_{l=1}^{L} w_l \tag{40}$$

Now we can eliminate $\lambda$ from (39) by (40) which leads to (32) for updating $\pi_k$.